%% file: main.tex
\pdfoutput=1

\documentclass[11pt]{article}

\usepackage{acl}

\usepackage{times}
\usepackage{latexsym}

\usepackage[T1]{fontenc}

\usepackage[utf8]{inputenc}

\usepackage{microtype}

\usepackage{times}
\usepackage{latexsym}
\usepackage{microtype}
\usepackage[LFE, LAE, T1]{fontenc}
\usepackage[utf8]{inputenc}
\usepackage{arabtex}
\vocalize 
\transtrue 
\arabtrue 
\usepackage{utf8}
\usepackage{arabtex}
\usepackage[russian, farsi, arabic, english]{babel}
\usepackage{CJKutf8}
\usepackage{amsmath}
\usepackage{times}
\usepackage{latexsym}
\usepackage{hyperref}       
\usepackage{url} 
\usepackage{booktabs} 
\usepackage{amsfonts} 
\usepackage{lipsum}
\usepackage{times}
\usepackage{latexsym}
\usepackage{caption}
\usepackage{times}
\usepackage{float}
\usepackage{refcount}
\usepackage{blindtext}
\usepackage{graphicx}
\usepackage{multirow}
\usepackage{enumitem}
\usepackage{enumitem}
\usepackage{graphicx, array, blindtext}
\usepackage{soul}
\usepackage{longtable}
\usepackage{gb4e}
\usepackage{tipa} 
\usepackage{setspace}%
\usepackage{makecell}
\noautomath

\title{Zero-shot Cross-Linguistic Learning of Event Semantics}

\author{
\makecell{
    Malihe Alikhani$^{1}$\unskip\enspace\enspace
    Thomas Kober$^{2}$\unskip\enspace\enspace 
    Bashar Alhafni\Thanks{{\quad Equal contribution.}}$^{\thickspace\thickspace 3}$\unskip\enspace\enspace
    Yue Chen$^{*4}$\unskip\enspace\enspace 
    Mert Inan$^{*1}$
    \\
    {\bf Elizabeth Nielsen}$^{*2}$\unskip\enspace\enspace 
    {\bf Shahab Raji}$^{*5}$\unskip\enspace\enspace 
    {\bf Mark Steedman}$^{2}$\unskip\enspace\enspace 
    {\bf Matthew Stone}$^{5}$
}
\\
\makecell{
    $^{1}$University of Pittsburgh\unskip\enspace\enspace
    $^{2}$University of Edinburgh\unskip\enspace\enspace
    $^{3}$New York University Abu Dhabi\unskip\enspace\enspace \\
    $^{4}$Indiana University\unskip\enspace\enspace
    $^{5}$Rutgers University\unskip\enspace\enspace \\
    \texttt{\{malihe,mert.inan\}@pitt.edu}, \texttt{\{tkober,steedman\}@inf.ed.ac.uk} \\
    \texttt{alhafni@nyu.edu}, \texttt{yc59@indiana.edu}, \texttt{e.k.nielsen@sms.ed.ac.uk}\\ \texttt{\{shahab.raji,mdstone\}@rutgers.edu} \\
}
}

\begin{document}
\maketitle
\begin{abstract}
Typologically diverse languages offer systems of lexical and grammatical aspect that allow speakers to focus on facets of event structure in ways that comport with the specific communicative setting and discourse constraints they face.  In this paper, we look specifically at captions of images across Arabic, Chinese, Farsi, German, Russian, and Turkish and describe a computational model for predicting lexical aspects. Despite the heterogeneity of these languages, and the salient invocation of distinctive linguistic resources across their caption corpora, speakers of these languages show surprising similarities in the ways they frame image content. We leverage this observation for zero-shot cross-lingual learning and show that lexical aspects can be predicted for a given language despite not having observed any annotated data for this language at all. 
\end{list}
\end{abstract}



%


 

\input{sections/intro.tex}
\input{sections/exp-design.tex}

\input{sections/analysis.tex}

\input{sections/ComputationalExperiments.tex}

\input{sections/conclusion.tex}

\input{sections/acknowledgements}
\bibliography{anthology, acl2021, custom}
\bibliographystyle{acl_natbib}

\section{Supplemental Material}
\label{sec:supplemental_a}
\input{sections/appendix_a.tex}

\end{document}

%% file: sections/intro.tex
\begin{figure*}[ht!]
\centering
\begin{tabular}{lr}
    \begin{minipage}{.32\textwidth}
        \includegraphics[height=3.9cm]{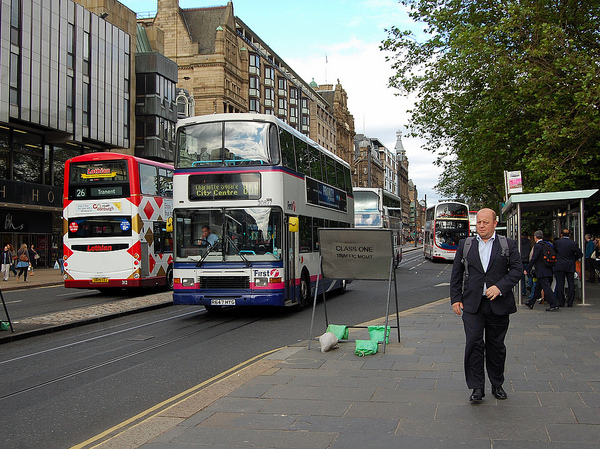}
    \end{minipage}
    &
        \begin{minipage}{.65\textwidth}
        \scalebox{0.8}{
        \begin{tabular}{lr}
        Arabic  &  
          \begin{tabular}[t]{rrrr}
          \AR{الطريق.}
          &\AR{بجانب}
          &\AR{يمشي}
          &\AR{رجل}
          \\
          street & nearby &  walking-\small{PRS}-\small{MASC}-\small{IPFV}-\small{3SG} & man
          \\
           \multicolumn{4}{r}{\small{A man is walking nearby the street.}}
          \end{tabular}
        
        \\ \\
        
        Chinese &  
          \begin{tabular}[t]{l}
          \begin{CJK}{UTF8}{bsmi}雙層 \space\space\space\space\space\space\space\space\space\space\space\space\space\space\space公共 \space汽車 正 \space在 \space\space\space\space公路 上 行駛\end{CJK}
          \\
          double-decker public bus now IPFV road\space\space on\space\space drive
          \\
          \multicolumn{1}{l}{\small{Double-decker public buses are driving on the road.}}
          \end{tabular}
          
        \\ \\
        
        Farsi   & 
          \begin{tabular}[t]{rrrrrr}
             \FR{می‌کنند.}
             &\FR{حرکت}
             &\FR{خیابان}
             &\FR{در}
             &\FR{دوطبقه}
             &\FR{اتوبوس‌های}
             \\
             do&move&street&in&double-decker&bus-PL
             \\
            \multicolumn{6}{r}{\small{Double-decker buses are moving in the street.}}  
          \end{tabular}
          
        \\ \\
          
        German & 
            \begin{tabular}[t]{lllllll}
            Zwei& 
            Busse & 
            fahren & 
            an &
            einer & 
            Haltelstelle & 
            vorbei.
            \\
            Two &buses&drive&&a&bus stop&past.
            \\
            \multicolumn{7}{l}{\small{Two buses drive past a bus stop.}}
            \end{tabular}
          %
          
         
         
         
            
        
            
            
        
        

        \end{tabular}
        }
        \end{minipage}

\end{tabular}
\caption{An example image from the MSCOCO dataset with Arabic, Chinese, German and Farsi captions. (ID: 000000568439, photo credit: Stephen Day)} 
\label{fig:examples}
\end{figure*}

\section{Introduction}

Tense and aspect rank among the most ubiquitous, problematic, and theoretically vexed features of natural language meaning \cite{sep-tense-aspect}.  Systems of tense and aspect differ considerably---but also often quite subtly---across languages. Figure \ref{fig:examples} shows how the corpus manifests differences and similarities across languages that align with their grammatical structures.
%
Tense and aspect have received extensive study across cognitive science; see \newcite{sep-tense-aspect}.  Nevertheless, from a computational point of view, it has been extremely challenging to gain empirical traction on key questions about them: how can we build models that ground speakers' choices of tense and aspect in real-world information?  how can we build models that link speakers' choices of tense and aspect to their communicative goals and the discourse context?  how can we build models that recognize tense and aspect?  
This is particularly challenging because we might have to work with small annotated datasets. 
The data scarcity
issue renders the need for effective cross-lingual
transfer strategies: how can one exploit abundant
labeled data from resource-rich languages to make
predictions in low resource languages? 

In this work, we leverage image descriptions to offer new insights into these questions. For the first time, we present a dataset of image descriptions and Wikipedia sentences annotated with lexical aspects in six languages. We hypothesize that across all of the languages that we study, image descriptions show strong preferences for specific tense, aspect, lexical aspect, and semantic field.  We adapt the crowdsourcing methodology used to collect English caption corpora such as MSCOCO and Flickr \cite{young2014image,lin2014microsoft} to create comparable corpora of Arabic, Chinese, Farsi, German, Russian, and Turkish image captions.   We extend the methodology of \newcite{alikhani2019caption} to get a synoptic view of tense, lexical aspect, and grammatical aspect in image descriptions in these diverse languages.

Finally, we study the extent to which verb aspect can be predicted from distributional semantic representations across different languages when the model was
never exposed to any data of the target language
during training, essentially performing zero-shot
cross-lingual transfer.
We consider predicting lexical aspect at the phrase level an important prerequisite for modelling fine grained entailment relations, such as inferring consequent states~\citep{Moens_1988}. For example, this is important for keeping knowledge bases up-to-date by inferring that the consequence of \emph{Microsoft \textbf{having acquired} GitHub}, is that now, \emph{Microsoft \textbf{owns} GitHub}.


Our results show that the grammatical structure of each language impacts how caption information is presented. 
Throughout our data, we find, as in Figure~\ref{fig:examples}, that captions report directly visible events, focusing on what's currently in progress rather than how those events must have begun or will culminate.
Yet they do so with different grammatical categories across languages: the progressive aspect of Arabic; the unmarked present of German; or the aspectual marker of the imperfective verbs of Chinese describing an event as in progress. 



\section{Related Work}

Linguists and computational linguists have largely focused on aspectuality as it has been used in unimodal communication. \newcite{caselli2007inferring} showed how aspectual information plays a crucial role in computational semantic and discourse analyses. \newcite{pustejovsky2010iso} described how aspect must be considered for event annotations and \newcite{baiamonte2016annotating} incorporated lexical aspect in the study of the rhetorical structure of text. \newcite{kober-etal-2020-aspectuality} presented a supervised model for studying aspectuality in unimodal scenarios only in English. In this work however, we focus on image captions that enable us to better understand how humans describe images. We also explore for the first time the potential of zero-shot models for learning lexical aspect across languages and genre.

The field of automatic image description saw an explosive growth
with the release of the Flickr30K and MSCOCO datasets \cite{vinyals2015show}. Fewer works however, have studied how humans produce image descriptions \cite{bernardi2016automatic,li2019visual}.
For example, \newcite{van-miltenburg-etal-2018-didec} studied the correlations between eye-gaze patterns and image descriptions in Dutch. \newcite{jas2015image} investigated the possibility of predicting
image specificity from eye-tracking data and \newcite{van-miltenburg-etal-2018-varying} discussed linguistics differences between written and spoken image descriptions. In this work we continue this effort by offering the first comparative study of verb use in image description corpora that we have put together in six different languages. 
\newcite{alikhani-etal-2020-cross,mccloud1993understanding,cohn2013visual,alikhani2018exploring,cumming2017conventions,alikhani2019cite} proposed that the intended contributions and inferences in multimodal discourse can be characterized as coherence relations. Our analyses and computational experiments explore the extent to which different grammatical-based distinctions correlate with discourse goals and contextual constraints and how these findings generalize across languages.





%% file: sections/exp-design.tex
\section{Data Collection and Annotation}
\label{sec:exp-design}


Given a set of images, subjects were requested to describe the images using the guideline that was used for collecting data for MSCOCO \cite{lin2014microsoft}. The instructions were translated to six target languages. For the Chinese instructions, we reduced the character limits from 100 to 20 since the average letter per word for English is 4.5. Generally, a concept that can be described in one word in English can also be described in one or two characters in Chinese. The original guideline in English as well as the translations can be found in the attached supplementary material.

We recruited participants through Amazon Mechanical Turk and Upwork.\footnote{\url{https://www.upwork.com/}} All subjects agreed to a consent form and were compensated at an estimated rate of USD 20 an hour.
We collected captions for 500 unique images (one caption per image in each of the languages that we study in this paper) that were randomly sampled from MSCOCO for each language. The results of our power analysis suggest that with this sample size, we are able detect effects sizes as small as 0.1650 in different distributions of lexical aspect with a significance level of 95\% \cite{faul2014g}.


\paragraph{Annotation Effort.}
The data is annotated by expert annotators for language specific characteristics of verbs such as tense, grammatical and lexical aspect and the Cohen Kappa inter-rater agreements \cite{cantor1996sample} are substantial ($\kappa > 0.8$) inter-annotator agreement across the languages. 

\subsection{Methods}
To compare captions and text in a different unimodal genre, we randomly selected 200 sentences across all languages from Wikipedia and annotated their lexical aspect. For Arabic, we used MADAMIRA \cite{pasha-etal-2014-madamira} to analyze the image captions which are written in Modern Standard Arabic.
%
We limited the 200 Chinese Wikipedia sentences to 20 characters in length. The word segmentation and part-of-speech tagging are performed using Jieba Python Chinese word segmentation module~\cite{sun2012jieba}. Traditional Chinese to Simplified Chinese character set conversion was done using zhconv.\footnote{\url{https://github.com/gumblex/zhconv}}
%

%
The Farsi image captions and the Wikipedia sentences were automatically parsed using \textit{Hazm} library.
%
For German, we used UDPipe~\citep{Straka_2017} and we have analysed the Russian morphological patterns by pymorphy2~\citep{Korobov_2015}.
For Turkish, the morphological analysis of all the verb phrases in the Wikipedia sentences and the captions are performed using the detailed analysis in \cite{Oflazer94anoutline}. While separating noun phrases from verb phrases, stative noun-verbs of existence (``var'' instead of ``var olmak'') were considered as verbs as well, following the analysis by ~\cite{cakmak2013}.
%



%% file: sections/analysis.tex
\section{Data Analysis}
\label{sec:analysis}
We performed an analysis of our data to study the following questions: What do image descriptions in Arabic, Chinese, Farsi, German, Russian and Turkish have in common? What are some of the language-specific properties? What opportunities do these languages provide for describing the content of images? 
In what follows, we first describe similarities across languages. Next we discuss languages specific properties related to tense and aspect.

In general, captions are less diverse as opposed to Wikipedia verb phrases in terms of their verbs vocabulary across the six languages.
Table~\ref{tab:topk_verbs} shows the accumulative percentage of top K verbs for the six languages for Wikipedia and image captions. Wikipedia sentences and captions have different distributions of tense, grammatical aspect and lexical aspect across all languages ($p<0.01$, $\chi>12.5$). When it comes to Arabic, atelic verbs dominate the verbs used in Arabic captions. However, the stative verbs dominate the verbs used in Wikipedia sentences.
%
%
\begin{table*}[ht!]
\centering
\begin{small}
\begin{tabular}{lcc|cc|cc|cc|cc|cc}
\toprule
      & \multicolumn{2}{c}{Arabic} & \multicolumn{2}{c}{Chinese} & \multicolumn{2}{c}{Farsi} & \multicolumn{2}{c}{German} & \multicolumn{2}{c}{Russian} &
      \multicolumn{2}{c}{Turkish}\\ \cmidrule{2-13} 
      & Wiki     & Capt.     & Wiki      & Capt.     & Wiki     & Capt. & Wiki & Capt. & Wiki & Capt. & Wiki & Capt.\\ 
      \midrule
\multicolumn{1}{l}{Top 10}  &   0.262       & 0.688          & 0.264          & 0.367       & 0.364         & 0.664  & 0.394 & 0.582  & 0.257 & 0.654  & 0.283 & 0.457 \\ 
\multicolumn{1}{l}{Top 30}  &   0.485       & 0.937          & 0.396          & 0.589       & 0.466         & 0.854  &   0.567   &   0.804  & 0.455 & 0.900  & 0.524 & 0.666   \\ 
\multicolumn{1}{l}{Top 100} &    0.832      & --          & 0.650          & 0.911       & 0.545         &  --   & 0.911 & --  & 0.802 & --  &  0.728 & 0.856  \\ 
\bottomrule
\end{tabular}
\end{small}
\caption{Captions show a limited distribution of verbs in comparison with Wikipedia. Verb use in Chinese and Turkish captions dataset are more diverse than in Farsi and Arabic caption datasets.}
\label{tab:topk_verbs}
\end{table*}
%

 
Moreover, present imperfective verbs make 99\% and present perfective verbs make 1\% of 85 inflected verbs across all Arabic captions. However, this is drastically different in our baseline. Across 200 full Arabic Wikipedia sentences and out of 180 inflected verbs, present perfective and present imperfective make 49.5\% and 2\% respectively. Whereas, past perfective and past imperfective make 44.6\% and 4\% respectively.

This largely agrees with what we analyzed for other languages. In the Chinese data, 56\% of Chinese caption verbs are imperfective whereas the majority (70\%) of the Chinese Wikipedia descriptions are stative. Chinese Wikipedia sentences also have very few atelic descriptions (1.8\%) whereas Chinese captions are populated with atelic descriptions. 
%
%
Chinese does not have tense, but we annotated the sentences both in captions and Wikipedia to learn about the number of sentences that present some kind of cues to refer to an event in the past i.e. adverb. In Wikipedia, 26\% of sentences refer to events in past but this number decreases to less than 1\% in captions. 
For Farsi, atelic events make up to 72\% of Farsi captions and 17\% of Farsi Wikipedia. As in Arabic and Chinese, we observed a major difference in distributions of grammatical aspect and tense in Farsi Wikipedia and Farsi captions. Farsi captions are populated with simple and imperfective present verbs. 
German captions also follow the general trend with 96\% of verbs in caption exhibiting imperfective aspect, in comparison to only 57\% in Wikipedia. 
Atelic verbs dominate the Aktionsart distribution of the captions dataset, making up 55\% of all verb occurrences, whereas only 16\% of verbs are atelic in the Wikipedia sample. The trend is conversed for telic verb occurrences, which make up only 4\% in the captions dataset, but 43\% in the Wikipedia sample. Interestingly, the proportion of stative verbs is roughly equal in captions and Wikipedia.

The Russian data also hold with these general trends: all captions are imperfective, whereas only 50\% of Wikipedia sentences are. This distribution is even more extreme in Russian than in other languages partially because of a unique property of the Russian aspectual and tense system: only verbs that refer to past or future events in Russian can be perfective. In the captions, 99\% of verbs refer to present events and therefore are required to be imperfective. This also is borne out the telicity of Russian captions: 49\% of captions are atelic, 30\% are stative, and only 22\% are telic. By contrast, only 21\% of Wikipedia data is atelic, while 26\% is stative, and 53\% is telic. As discussed in Section \ref{sec:specific} below, this reflects a correlation between perfectivity and telicity in Russian.

Telicity of the Turkish data follows a similar distribution to the other languages, with a key difference in the statistics of stative verbs. Both Wikipedia sentences and captions have higher count of stative verbs compared to other languages. 56\% of Wikipedia verbs and 63\% of caption verbs are stative in Turkish. This is caused by the inherent copula usage and preference of stative and timeless tenses such as the ``geniş zaman''. Atelic verb percentage in captions (30.4\%) is considerably smaller to that of stative verbs (63.8\%). There is a drastic difference between the number of telic verbs with a 32.4\% in Wikipedia phrases compared to 5.8\% in captions.

\begin{table*}[ht!]
\centering
\begin{small}
\begin{tabular}{lcc|cc|cc|cc|cc|cc}
\toprule
                              & \multicolumn{2}{c}{Arabic} & \multicolumn{2}{c}{Chinese} & \multicolumn{2}{c}{Farsi} & \multicolumn{2}{c}{German} & \multicolumn{2}{c}{Russian} &
                              \multicolumn{2}{c}{Turkish} \\ \cmidrule{2-13} 
                              & Wiki     & Capt.     & Wiki      & Capt.     & Wiki     & Capt.    & Wiki & Capt.     & Wiki & Capt.  & Wiki & Capt.  \\ \midrule
Atelic  & 0.089         & 0.722       & 0.018          & 0.561       & 0.171         & 0.719   & 0.162 & 0.550  & 0.213 &  0.488 & 0.114 & 0.304  \\ 
Telic   & 0.371         & 0.010       & 0.285          & 0.063       & 0.470         & 0.042    &  0.431   & 0.038 & 0.530 & 0.218 & 0.324 & 0.058\\ 
Stative & 0.540         & 0.268       & 0.698          & 0.377       & 0.357         & 0.237    &  0.407   &   0.412 & 0.257 & 0.299 & 0.560 & 0.638 \\ 
\bottomrule
\end{tabular}
\end{small}
\caption{Captions include more atelic descriptions in comparison with Wikipedia across languages.}
\label{tab:aktionsart}
\end{table*}


\subsection{Language-Specific Observations} \label{sec:specific}
\paragraph{Arabic.} 

Arabic has a rich morphological system \cite{arabic-nlp-intro}.
Moreover, verbs in Arabic have three grammatical aspects: perfective, imperfective, and imperative. The perfective aspect indicates that actions described are completed as opposed to the imperfective aspect which does not specify any such information. Whereas the imperative aspect is the command form of the verb. 

Similar to German and Russian, non-past imperfective verbs were dominant across the captions in Arabic as opposed to Chinese, Farsi, and Turkish. 
%
Furthermore and as shown in Table~\ref{tab:aktionsart}, 72.2\% of Arabic captions were atelic, and this is the highest atelic percentage for captions across all languages. Whereas, 8.9\% of the Arabic Wikipedia sentences were atelic, which constitutes the lowest atelic percentage for Wikipedia sentences across all other languages. This highlights an interesting evidence of the morphological richness in Arabic and how verbs can inflect for mood and aspect.

\paragraph{Chinese.} 
Chinese is an equipollent-framed language (E-framed language), due to its prominent feature -- serial verb construction~\cite{slobin2004many}. For example,
\begin{CJK}{UTF8}{gbsn}
走进 (walk into) and 走出 (walk out of)
\end{CJK}
are treated as two different verbs.
This phenomenon greatly enlarged the vocabulary of Chinese verbs perceived by POS taggers and parsers. We believe this is an important reason why Chinese verbs look so diverse and the distribution among atelic, telic and stative looks rather imbalanced. Having the base verb character and adding on aspectual particles changes the telicity. Given the nature of Wikipedia text, it is observed that in table~\ref{tab:aktionsart} only 1.8\% are atelic and more than 69.8\% are stative, while in image captions more than 56\% are atelic.

Since Chinese does not have the grammatical category of tense, the concept denoted by tense in other languages is indicated by content words like adverbs of time or it is simply implied by context.  For example, the verb for ``do'' is 
\begin{CJK}{UTF8}{gbsn}
做 (zuo)
\end{CJK}, which is used to describe all past, present, and future events. Since the verb remains the same, temporal reference is instead indicated by the time expressions~\cite{lin2006time}, for example: 
\begin{exe}
    \ex \begin{CJK}{UTF8}{gbsn}昨天\space\space\space\space\space\space\space我做了\space\space\space\space\space批萨。\end{CJK}
    \\
    Yesterday I do\space PFV pizza.\\
    Yesterday I made pizza.
\end{exe}

\paragraph{Farsi.}

In the Farsi caption dataset four verbs make up to around 50\% of the verbs: 
\textit{to be} 
(\FR{بودن}),
\textit{to play} 
(\FR{بازی کردن})
, \textit{to sit} 
(\FR{نشستن}),
and \textit{to look}
(\FR{نگاه کردن})

%
Table ~\ref{tab:topk_verbs} shows difference in verbs distributions across languages. The data regarding the distribution of caption verbs in English are reported by \cite{alikhani2019caption}. Chinese captions are much more diverse and the difference is statistically significant ($p < 0.05$, $\chi = 14.4$).


Farsi verbs are either simple or compound. 
Any lexical unit which contains only a verbal root is a simple verb (e.g. verbal root: \FR{رفتن} ‘to go'). The lexical unit which contain either a prefix plus a verbal root, or a nominal plus either a regular verbal root or an auxiliary verb are compound verbs. Related to this is the phenomenon of incorporation, defined by
\cite{spencer1991morphological} as the situation in which “a word forms a kind of compound with its direct object, or adverbial modifiers while retaining its original syntactic function.”




59.3\% of Farsi Caption verbs are compound and 88.2\% of the compound verbs are constructed with 
\FR{کردن}
(to do) and
\FR{شدن}
(to be). Wikipedia on the other hand includes only 12.1\% compound verbs. \newcite{Nemood} conjectured that
\FR{کردن}
(to do) and
\FR{شدن}
(to be) are used when the speaker wants to highlight the meaning of the noun even more in comparison with cases where nouns are accompanied with
\FR{گرفتن} 
(to take) or
\FR{داشتن}
(to have).
For example,
\FR{نگاه کردن}
(literally \textit{Do a look}) is the fourth most frequent verb in captions.

However, the majority (97\%) of the compound verbs in captions are constructed with nouns. 

\newcite{megerdoomian2002aspect} hypothesized that the aspectual properties depend on the interaction between the non-verbal and the light verb and that the choice of light verb affects argument structure. For instance, to form the transitive version of an intransitive predicate, Farsi speakers replace the light verb by its causative form. \textbf{All of the intransitive compound verbs in our corpus are atelic.} 
\paragraph{German.}

German speakers predominantly used the present simple --- rather than the present progressive --- to describe atelic activities, where we found that only $\approx$7\% of atelic captions have been described in the present progressive. For example, sentences~\ref{ex:play_wii}-\ref{ex:skiing} below show two captions where the ongoing activity is described in the present simple in German, however in English, the present progressive would be used. In English, the use of the present simple has a strong futurate reading, which is substantially weaker in German. Thus we attribute the frequent use of the present simple in German to it being less aspectually ambiguous.

\begin{enumerate}[label={(\arabic*)}, resume*]
\itemsep0em
\item \label{ex:play_wii} Zwei Männer \textbf{spielen} Wii im Wohnzimmer.
\item[] \emph{Two men \textbf{are playing} on a Wii in the living room}.
\item \label{ex:skiing} Ein Mann und eine Frau \textbf{fahren} Ski.
\item[] \emph{A man and a woman \textbf{are skiing}}.
\end{enumerate}

We furthermore found that German speakers have frequently omitted the verb altogether if an imaged depicted some form of still life. These sentences exhibit stative lexical aspect, and typically, verbs such as ``stand", ``lie" or a form of ``to be" would have been the correct verb 
as sentences~\ref{ex:zug_steht}-\ref{ex:pizza_beer}
below demonstrate, where we have added a plausible verb in square brackets.

\begin{enumerate}[label={(\arabic*)}, resume*]
\itemsep0em
\item \label{ex:zug_steht} Ein Zug \textbf{[steht]} neben einer Ladeplattform.
\item[] \emph{A train \textbf{[is standing]} next to a loading bay}.
\item \label{ex:pizza_beer} Eine Pepperoni Pizza \textbf{[liegt]} in einer Pfanne neben einem Bier.
\item[] \emph{A pepperoni pizza \textbf{[is lying]} in a pan next to a beer}.
\end{enumerate}

\paragraph{Russian.} 
A distinction between imperfective and perfective aspect must be marked on all Russian verbs. This contrasts with languages (e.g., Spanish) where aspect is only marked explicitly in a subset of the verbal system, such as within the past tense. Aspect marking in Russian is often done by means of affixation: a default-imperfective stem becomes perfective with the addition of a prefix (e.g.\ \textit{pisat'} > \textit{napisat'} `to write' \cite{Laleko_2008}). 
Perfective aspect expresses a view of an event ``in its entirety'' \cite{comrie1976aspect}, including its end point, meaning that perfectivity and telicity are highly correlated. For example, the use of the perfective \textit{napisat'} `to write' implies the completion of a finite amount of writing, whether or not the speaker chooses to include an explicit direct object indicating what is being written. There is disagreement in the literature on whether all perfective verbs in Russian are telic or if the perfectivity is merely correlated with telicity \cite{Gueron_2008,Filip_2004}. However, the fact that all verbs must be explicitly marked as either perfective or imperfective, combined with the fact that telicity is at least positively correlated with perfectivity, may lead to more verbs in the Russian being labelled as telic. In fact, we do find that when compared with languages such as English, where verbs may remain under-specified for aspect and therefore for telicity, the Russian captions contain significantly more telic verbs.

\paragraph{Turkish.}
Lexical aspects of verbs in Turkish captions differ from other languages in terms of choice of the sentence structure and the diversity of Turkish tenses, with the presence of copula. These intricacies are analyzed using the work of \cite{dilder602689} and \cite{yesim2003}. It can be observed that Turkish-speakers tend to choose a specific sentence structure while describing pictures. 

Captions are populated with noun phrases consisting of a verbal adjective, a subject and an implicit noun-verb (``var''). 
%
%
The most important aspect about determining lexical aspect in Turkish is the plethora of tenses.
A considerably different tense is the ``geniş zaman'', which translates to "broad time/tense". Its use broadens the time aspect in a verb to an extent that the verb exists in a timeless space. Even though it is generally compared with the present simple tense in English, ``geniş zaman'' telicity greatly depends on the context and the preceding tense in the agglutinative verb structure. Wikipedia sentences contain 13.3\% ``geniş zaman'' verbs while caption verbs do not have any of that formation. This is due to the difference of giving a description or a definition. 

Turkish definitions are timeless and use ``geniş zaman'' more frequently, while descriptions, like in the captions, use other tenses. It can be presumed that all ``geniş zaman'' verbs are atelic; however, this does not necessarily hold true in captions where a limited number of telic cases exist, which increases the importance of a differentiation between atelic and telic tenses in Turkish. 
Another distinction that is visible between the Turkish image captions and Wikipedia sentences is the progressive aspect. 59.7\% of caption verbs are progressive while only 0.9\% of Wikipedia verbs are progressive. This aspect is used extensively in captions due to its close relation with any action verb that is being done.


%% file: sections/ComputationalExperiments.tex
\begin{table*}[!t]
\begin{tabular}{clcc|cc|cc|cc|cc|cc}
\toprule
&\multicolumn{1}{c}{\multirow{2}{*}{\textbf{Aspect}}}& \multicolumn{2}{c}{Arabic}  & \multicolumn{2}{c}{Chinese}   & \multicolumn{2}{c}{Farsi} & \multicolumn{2}{c}{German} & \multicolumn{2}{c}{Russian} & \multicolumn{2}{c}{Turkish}\\ 
\cmidrule{3-14} 
&\multicolumn{1}{c}{} & Capt. & Wiki & Capt. & Wiki & Capt. & Wiki & Capt. & Wiki & Capt. & Wiki & Capt. & Wiki \\ \midrule
\multirow{3}{*}{\rotatebox[origin=c]{90}{\textbf{fastText}}}& Atelic & \textbf{0.95} & - & \textbf{0.97} & - & \textbf{0.95} & - & \textbf{0.90} & - & \textbf{0.96} & - & 0.51 & - \\ 
& Telic & - & 0.48 & - & 0.00& - & 0.74 & - & \textbf{0.89} & - & 0.83 & - & 0.62\\
& State & 0.84 & 0.66 & 0.00& \textbf{0.89} & 0.83 & 0.59 & \textbf{0.88} & \textbf{0.88} & \textbf{0.94} & 0.27 & 0.83 & 0.80\\\bottomrule

\multirow{3}{*}{\rotatebox[origin=c]{90}{\textbf{mBERT}}} & Atelic & 0.50 & - & 0.80 & - & 0.73 & - & 0.72 & - & 0.78 & - & \textbf{0.96} & - \\ 
& Telic & - & 0.64 & - & \textbf{0.92}& - & 0.75 & - & 0.84 & - & \textbf{0.83} & - & \textbf{0.79}\\
& State & 0.88 & \textbf{0.79} & \textbf{0.91}& 0.47 & \textbf{0.93} & 0.57 & 0.82 & 0.82 & 0.88 & \textbf{0.44} & 0.91 & \textbf{0.89}\\\bottomrule
\multirow{3}{*}{\rotatebox[origin=c]{90}{\textbf{ELMo}}} & Atelic & 0.65 & - & 0.76 & - & 0.77 & - & 0.78 & - & 0.90 & - & \textbf{0.97} & - \\ 
& Telic & - & \textbf{0.66} & - & 0.87 & - & \textbf{0.79} & - & 0.76 & - & \textbf{0.83} & - & 0.74\\
& State & \textbf{0.89} & 0.78 & 0.88 & 0.22 & \textbf{0.93} & \textbf{0.67} & 0.85 & 0.75 & \textbf{0.94} & 0.20 & \textbf{0.93} & 0.86\\\bottomrule

\end{tabular}
\caption{Mono-lingual F1-scores per label across all languages with using fastText embeddings (top), multilingual BERT embeddings (middle) and ELMo embeddings (bottom).}
\label{tab:mono_lingual}
\end{table*}

\begin{figure*}[!htb]
\centering
\includegraphics[width=\textwidth]{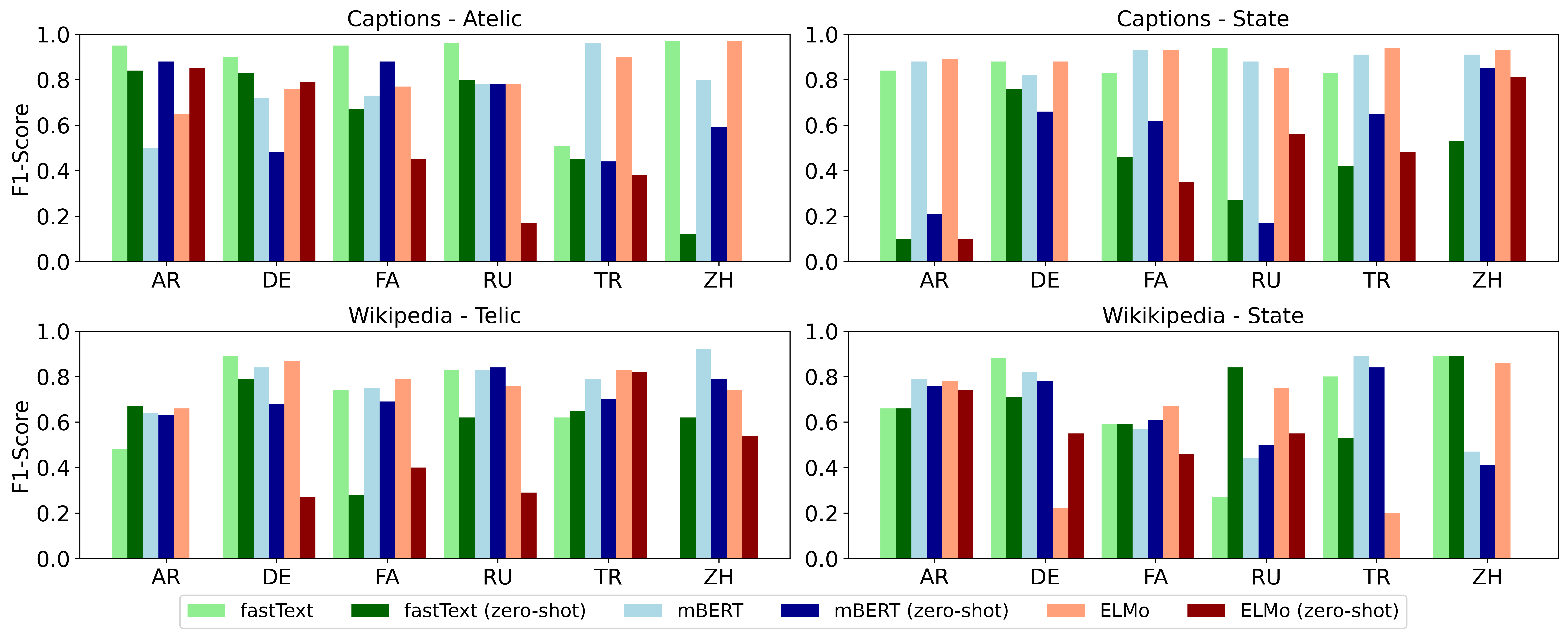}
\captionsetup{font=small}
\caption{Performance comparison between zero-shot cross-lingual (darker shades) learning and a mono-lingual (lighter shades) setup. Remarkably, even without any target language data, our simple zero-shot setup is competitive with using mono-lingual data and even surpasses it in some cases.}
\label{fig:zero_shot_cross_lingual}
\end{figure*}

\section{Computational Experiments}

In this section we leverage our multilingual annotated dataset and investigate to what extent aspect can be detected with computational methods. More specifically, the primary research question we address in this section is an empirical investigation whether distributional semantic models capture enough information about the latent semantics of aspect to be detected across languages.

Our use of distributional semantic representations is furthermore motivated by the fact that they are readily available in numerous languages, and that they, contrary to manually constructed lexicons such as VerbNet~\citep{Schuler_2005} or LCS~\citep{Dorr_1997}, scale well with growing amounts of data and across different languages. Furthermore, there is a growing body of evidence that models based on the distributional hypothesis capture some facets of aspect~\citep{kober-etal-2020-aspectuality,metheniti-etal-2022-time}, despite the fact that aspect is represented in a very diverse manner across languages.  

\subsection{Aspectual Classification}

We treat the prediction of verb aspect as a supervised classification task and experiment with pre-trained fastText~\citep{Grave_2018} embeddings\footnote{\url{https://fasttext.cc/docs/en/pretrained-vectors.html}}, multilingual BERT~\citep{devlin2019bert}, and ELMo~\citep{peters-etal-2018-deep,che-EtAl:2018:K18-2}\footnote{While the BERT model is truly multilingual, we use a single monolingual ELMo model for our experiments from~\url{https://github.com/HIT-SCIR/ELMoForManyLangs}.} as input, and the aspectual classes \emph{state, telic, atelic} as targets. For fastText we average the word embeddings to create a single vector representation, for multilingual BERT we use its \texttt{[CLS]} token, and for ELMo the pooled representation of the encoded utterance for classification. We use the Logistic Regression classifier from scikit-learn~\citep{Pedregosa_2011} with default hyperparameter settings. 

Our choice of models is motivated by: a) assessing performance with a word-level model (fastText), b) estimating the performance difference when large pre-trained models (ELMo \& mBERT) are applied, and c) observing the difference between a single multilingual model (mBERT) and monolingual models for the different languages (fastText \& ELMo).

\paragraph{Mono-lingual.} For the mono-lingual experiments, we evaluate our method on the annotated captions and Wikipedia sentences, however we decided to drop all \emph{telic} instances from the captions data, and all \emph{atelic} instances the Wikipedia sentences, as they occur very infrequently in either respective corpus.\footnote{This reduced the classification problem to a 2-class problem, \emph{stative} vs. \emph{non-stative}.} We are focused on establishing whether aspect can be predicted from embeddings across languages \emph{in principle} and wanted to avoid obfuscating the problem of predicting aspect with the problem of class imbalance. 

The aim of our first experiment is to establish that aspect can be classified for our set of languages with distributional representations in a supervised setting as has been shown on English  data~\citep{kober-etal-2020-aspectuality}.
\begin{figure}[!htb]
\centering
\includegraphics[width=\columnwidth]{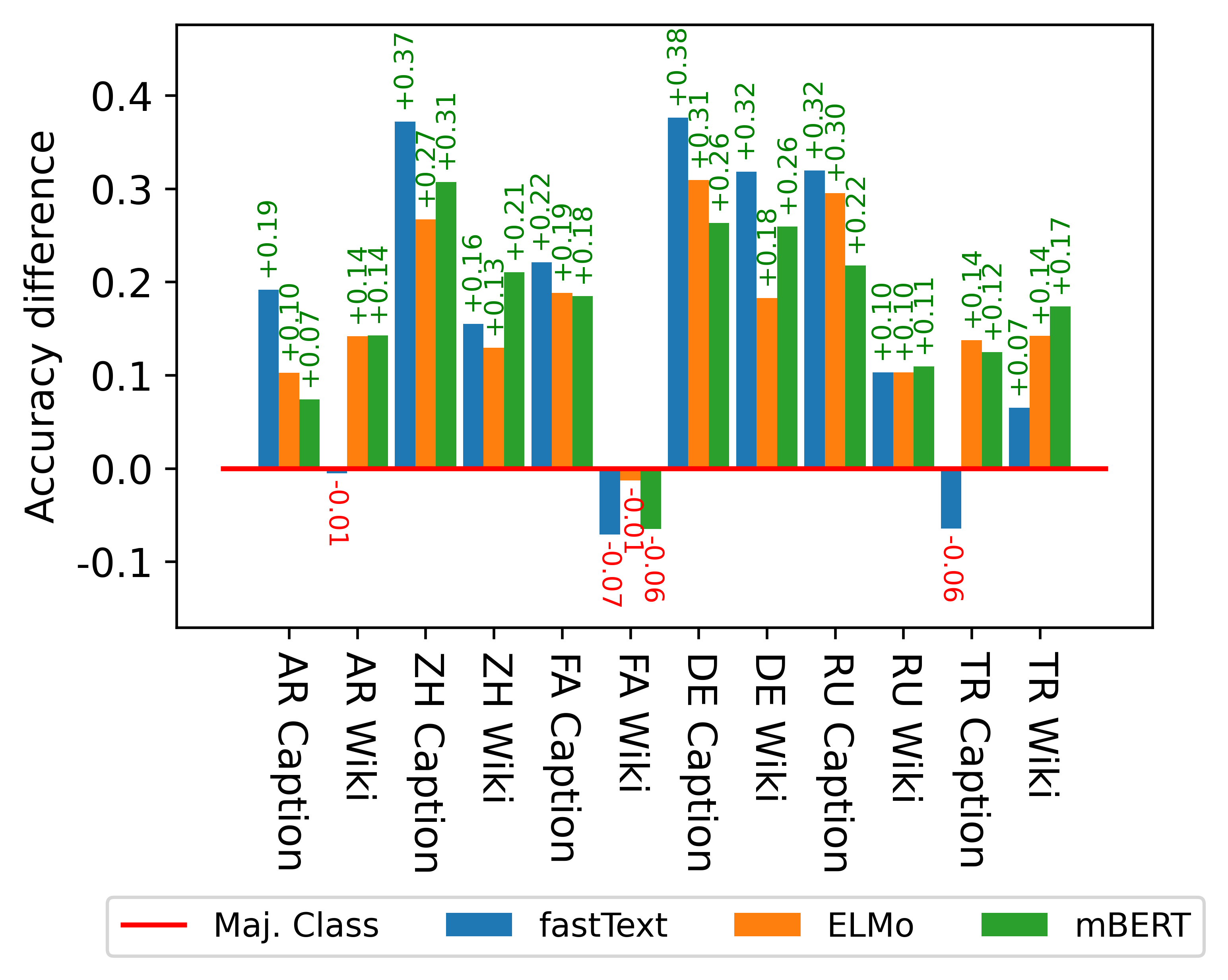}
\captionsetup{font=small}
\caption{Accuracy comparison of a majority class baseline to fastText, multilingual BERT and ELMo models across all our target languages and domains.}
\label{fig:monolingual_baseline}
\end{figure}
Figure~\ref{fig:monolingual_baseline} shows the difference in Accuracy of our models in comparison to a majority class baseline. As the figure shows, the distributional models are able to outperform the majority class baseline by substantial margins across the board with the exception of our Farsi Wikipedia dataset where we underperform the baseline by a small margin.

Next, we aim to establish baseline scores for the distributional models on our dataset. We perform stratified 10-fold cross-validation on our annotated datasets and report a micro-averaged F1-Score on the basis of accumulating the number of true positives, false positive, and false negatives across all cross-validation runs~\citep{Forman_2010}.

Table~\ref{tab:mono_lingual} shows that except for Chinese, our simple method of predicting aspect from averaged fastText embeddings works astonishingly well across languages, achieving F1-scores in the mid-80s to mid-90s for many languages. Multilingual BERT and ELMo perform similarly across languages with notable problems for distinguishing between states and \emph{telic} events in Russian and Chinese.

Overall, all models perform approximately in the same ballpark, specifically, there is no dramatic loss in performance when using a single multilingual model in comparison to monolingual models. Conversely, an LSTM-based model and the even simpler bag-of-words based model work remarkably well given the latent nature of aspect.  Distributional representations appear to capture enough information for making fine-grained semantic distinctions --- an important result for further work on multilingual semantic inference around consequence and causation~\citep{mirza-tonelli-2014-analysis,kober-etal-2019-temporal,guillou-etal-2020-incorporating}.

\paragraph{Zero-Shot Cross-lingual.} For the zero-shot cross-lingual experiment we use the aligned fastText embeddings and the same mBERT and ELMo models as in the mono-lingual experiments.\footnote{\url{https://fasttext.cc/docs/en/aligned-vectors.html}} We perform a zero-shot learning on the basis of a  leave-one-language-out evaluation. This means that we train our Logistic Regression classifier on the data of five languages and evaluate performance on the sixth one. The models were never exposed to \emph{any} data of the target language during training, thereby performing zero-shot cross-lingual transfer. This assesses how much information can be leveraged cross-lingually, which has potential further applications for transfer learning and data augmentation. 

As for the mono-lingual experiments we drop the \emph{telic} class from the captions data, and the \emph{atelic} class from the Wikipedia data. Figure~\ref{fig:zero_shot_cross_lingual} compares mono-lingual with zero-shot cross-lingual performance, showing that our simple setup yields remarkably strong results, that in some cases even outperform the mono-lingual setup. Our results indicate that a considerable amount of aspectual information can be transferred and induced cross-lingually, providing a very promising avenue for future work.\footnote{A multilingual companion table to Table~\ref{tab:mono_lingual} is presented in Table~\ref{tab:zero_shot_cross_lingual_result_table} in Appendix~\ref{sec:supplemental_a}.}
\begin{figure*}[!htb]
\centering
\includegraphics[width=\textwidth]{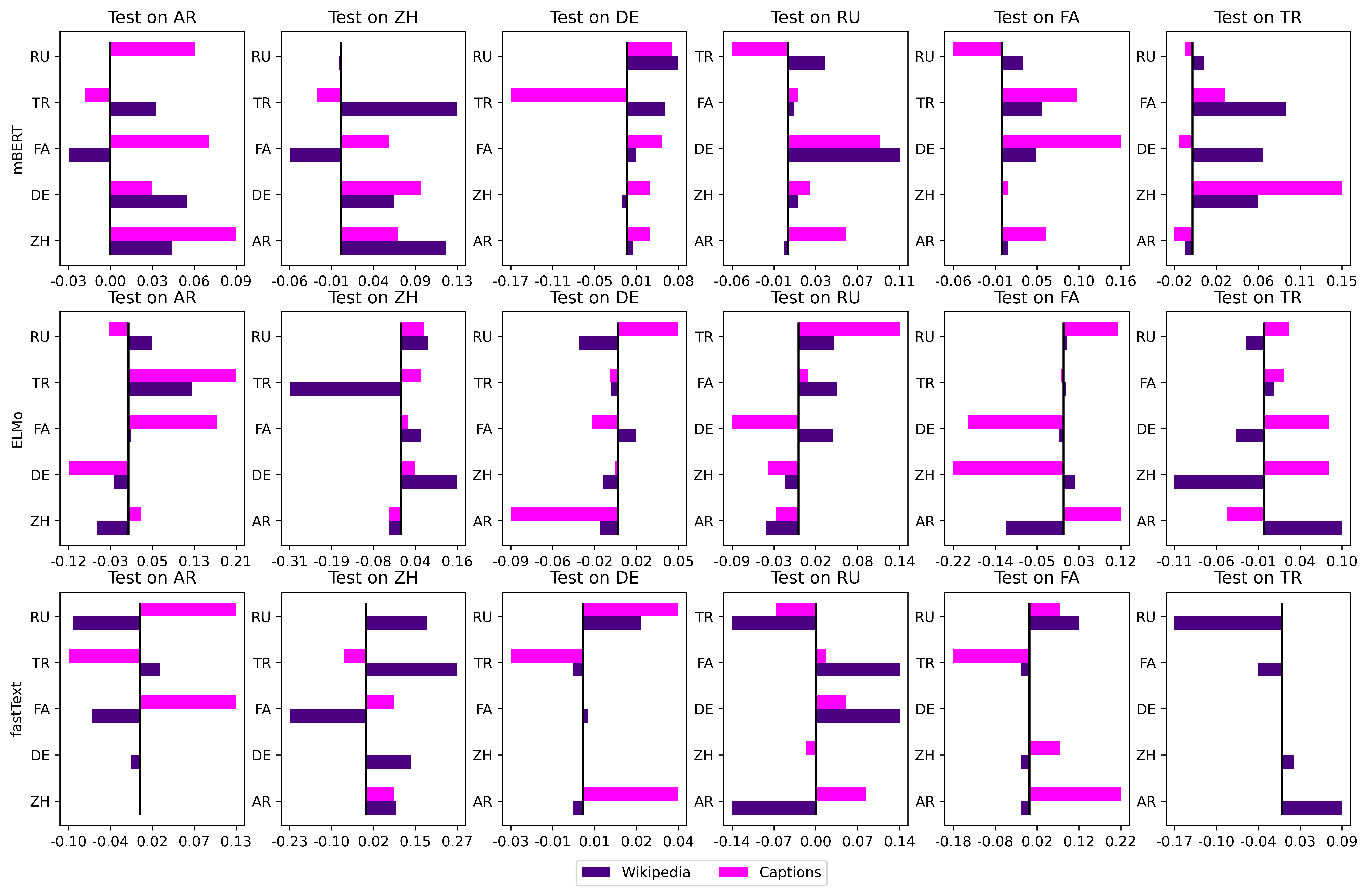}
\captionsetup{font=small}
\caption{Shapely-flavoured analysis of the impact of each language's presence in the training data on predicting aspect in a target language in a zero-shot cross-lingual setting.}
\label{fig:zero_shot_cross_lingual_shap}
\end{figure*}
In order to estimate the importance of the contribution of each language in the zero-shot setting we conduct a Shapely-flavoured~\citep{Shapely_1953} analysis. Shapely values are a method for quantifying the contribution to model performance of any given feature in a dataset~\citep{molnar2022}. 

As Shapely values operate on the \emph{feature} space, rather than the \emph{instance} space, we interpret the presence of training data for a particular languages as a binary indicator feature. This means that any languages can be ``active'' during model training, or not. This way, we can observe the performance of a model with and without any given language in the training data, and estimate that language's impact on model performance. 
The process to estimate the Shapely-flavoured impact value for a given language is perhaps best explained by an example: supposing our target language --- for which we want to predict aspect --- is Arabic, and we want to quantify the contribution of German language training data in our model, we start by training a model on Farsi data and compare our model's predictive performance to a model trained on Farsi \emph{and} German data. Next, we train our model on Farsi and Russian data, and compare its performance to a model trained on Farsi, Russian \emph{and} German data, and so on for all combinations of training data. Lastly, we average the differences of all these comparisons to obtain a value that represents the impact of German data on predicting aspect for Arabic. We perform this method for all model, language and domain combinations, with the resulting Figure~\ref{fig:zero_shot_cross_lingual_shap} summarising all Shapely-flavoured impact values for all languages. The figure shows the positive and negative impact of each language --- for the captions dataset in magenta and the Wikipedia dataset in indigo --- for measuring accuracy.  Generally, the impact of each language on model performance is primarily governed by the \emph{kind of model}, rather than the language(s) used for training. While this may seem somewhat dissatisfying at first, we believe that understanding model behaviour is paramount for transfer learning with cross-lingual data with the goal of leveraging e.g. the explicit aspectual markers in the Slavic languages to learn models for languages such as English where aspect is more opaque, as a very fruitful avenue for future research.

%% file: sections/conclusion.tex
\section{Conclusion}
\label{sec:dic}

By analyzing verb usage in image–caption corpora in Arabic, Chinese, Farsi, German, Russian and Turkish we find that people describe
visible eventualities as continuing and indefinite in temporal extent. We show that distributional semantic can reliably predict aspectual classes across languages, and achieves remarkable performance even in zero-shot cross-lingual experiments.

Our study has also revealed that these qualitative properties and grammatical differences reflect the discourse constraints in play when
subjects write captions for images and that these findings are generalizable across languages. 
We have leveraged this observation for our computational work where we show that aspect can be predicted with distributional representations in a mono-lingual setup. We have furthermore provided first evidence that aspect can be predicted in a zero-shot cross-lingual manner where a model has not been exposed to any training data in the target language at all.

%% file: sections/acknowledgements.tex
\section*{Acknowledgement}
We would like to thank Aaron White, Gabriel Greenberg and the anonymous reviewers for their helpful comments. The research presented here is supported by
NSF Awards IIS-1526723 and CCF-1934924.

%% file: sections/appendix_a.tex
\begin{table}[h!]
\centering
\begin{tabular}{lcc}
\toprule
         & Wikipedia     & Caption\\ \midrule
Arabic   & 11.60  &  4.65\\ 
Chinese  & 21.13  & 10.63 \\ 
Farsi    & 24  & 7 \\
German   & 13.43  &  9.47 \\
Russian  & 15.43  &  4.27 \\
Turkish  & 12.76 & 10.90 \\
\bottomrule
\end{tabular}
\caption{ Wikipedia sentences are on average longer, i.e. contain more tokens, than captions.}
\label{tab:sent_len_crossling}
\end{table} 




\begin{table*}[!th]
\begin{tabular}{clcc|cc|cc|cc|cc|cc}
\toprule
&\multicolumn{1}{c}{\multirow{2}{*}{\textbf{Aspect}}}& \multicolumn{2}{c}{Arabic}  & \multicolumn{2}{c}{Chinese}   & \multicolumn{2}{c}{Farsi} & \multicolumn{2}{c}{German} & \multicolumn{2}{c}{Russian} & \multicolumn{2}{c}{Turkish}\\ 
\cmidrule{3-14} 
&\multicolumn{1}{c}{} & Capt. & Wiki & Capt. & Wiki & Capt. & Wiki & Capt. & Wiki & Capt. & Wiki & Capt. & Wiki \\ \midrule
\multirow{3}{*}{\rotatebox[origin=c]{90}{\textbf{fastText}}}& Atelic & 0.84 & - & 0.12 & - & 0.67 & - & \textbf{0.83} & - & \textbf{0.80} & - & \textbf{0.45} & - \\ 
& Telic & - & \textbf{0.67} & - & 0.62& - & 0.28 & - & \textbf{0.79} & - & 0.62 & - & 0.65\\
& State & 0.10 & 0.66 & 0.53& \textbf{0.89} & 0.46 & 0.59 & \textbf{0.76} & 0.71 & 0.27 & \textbf{0.84} & 0.42 & 0.53\\\bottomrule

\multirow{3}{*}{\rotatebox[origin=c]{90}{\textbf{mBERT}}} & Atelic & \textbf{0.88} & - & \textbf{0.59} & - & \textbf{0.88} & - & 0.48 & - & 0.78 & - & 0.44 & - \\ 
& Telic & - & 0.63 & - & \textbf{0.79}& - & \textbf{0.69} & - & 0.68 & - & \textbf{0.84} & - & 0.70\\
& State & \textbf{0.21} & \textbf{0.76} & \textbf{0.85}& 0.41 & \textbf{0.62} & \textbf{0.61} & 0.66 & \textbf{0.78} & 0.17 & 0.50 & \textbf{0.65} & \textbf{0.84}\\\bottomrule
\multirow{3}{*}{\rotatebox[origin=c]{90}{\textbf{ELMo}}} & Atelic & 0.85 & - & 0.00 & - & 0.45 & - & 0.79 & - & 0.17 & - & 0.38 & - \\ 
& Telic & - & 0.00 & - & 0.54 & - & 0.40 & - & 0.27 & - & 0.29 & - & \textbf{0.82}\\
& State & 0.10 & 0.74 & 0.81 & 0.00 & 0.35 & 0.46 & 0.00 & 0.55 & \textbf{0.56} & 0.55 & 0.48 & 0.00\\\bottomrule

\end{tabular}
\caption{Zero-shot cross-lingual F1-scores per label across all languages with using fastText embeddings (top), multilingual BERT embeddings (middle) and ELMo embeddings (bottom).}
\label{tab:zero_shot_cross_lingual_result_table}
\end{table*}

%% file: main.bbl
\begin{thebibliography}{55}
\expandafter\ifx\csname natexlab\endcsname\relax\def\natexlab#1{#1}\fi

\bibitem[{Aksan and Aksan(2006)}]{dilder602689}
Mustafa Aksan and Yeşim Aksan. 2006.
\newblock Denominal verbs and their aspectual properties.
\newblock \emph{Dil Dergisi}, pages 7 -- 27.

\bibitem[{Aksan(2003)}]{yesim2003}
Yeşim Aksan. 2003.
\newblock \href {http://dad.boun.edu.tr/en/download/article-file/224716}
  {Türkçe'de durum değişikliği eylemlerinin kılınış özellikleri.}
\newblock \emph{DİLBİLİM ARAŞTIRMALARI DERGİSİ}.

\bibitem[{Alikhani et~al.(2019)Alikhani, Chowdhury, de~Melo, and
  Stone}]{alikhani2019cite}
Malihe Alikhani, Sreyasi~Nag Chowdhury, Gerard de~Melo, and Matthew Stone.
  2019.
\newblock Cite: A corpus of image-text discourse relations.
\newblock \emph{arXiv preprint arXiv:1904.06286}.

\bibitem[{Alikhani et~al.(2020)Alikhani, Sharma, Li, Soricut, and
  Stone}]{alikhani-etal-2020-cross}
Malihe Alikhani, Piyush Sharma, Shengjie Li, Radu Soricut, and Matthew Stone.
  2020.
\newblock \href {https://doi.org/10.18653/v1/2020.acl-main.583} {Cross-modal
  coherence modeling for caption generation}.
\newblock In \emph{Proceedings of the 58th Annual Meeting of the Association
  for Computational Linguistics}, pages 6525--6535, Online. Association for
  Computational Linguistics.

\bibitem[{Alikhani and Stone(2018)}]{alikhani2018exploring}
Malihe Alikhani and Matthew Stone. 2018.
\newblock Exploring coherence in visual explanations.
\newblock In \emph{2018 IEEE Conference on Multimedia Information Processing
  and Retrieval (MIPR)}, pages 272--277. IEEE.

\bibitem[{Alikhani and Stone(2019)}]{alikhani2019caption}
Malihe Alikhani and Matthew Stone. 2019.
\newblock “caption” as a coherence relation: Evidence and implications.
\newblock In \emph{Proceedings of the Second Workshop on Shortcomings in Vision
  and Language}, pages 58--67.

\bibitem[{Baiamonte et~al.(2016)Baiamonte, Caselli, and
  Prodanof}]{baiamonte2016annotating}
Daniela Baiamonte, Tommaso Caselli, and Irina Prodanof. 2016.
\newblock Annotating content zones in news articles.
\newblock \emph{CLiC it}, page~40.

\bibitem[{Bernardi et~al.(2016)Bernardi, Cakici, Elliott, Erdem, Erdem,
  Ikizler-Cinbis, Keller, Muscat, and Plank}]{bernardi2016automatic}
Raffaella Bernardi, Ruket Cakici, Desmond Elliott, Aykut Erdem, Erkut Erdem,
  Nazli Ikizler-Cinbis, Frank Keller, Adrian Muscat, and Barbara Plank. 2016.
\newblock Automatic description generation from images: A survey of models,
  datasets, and evaluation measures.
\newblock \emph{Journal of Artificial Intelligence Research}, 55:409--442.

\bibitem[{Cantor(1996)}]{cantor1996sample}
Alan~B Cantor. 1996.
\newblock Sample-size calculations for cohen's kappa.
\newblock \emph{Psychological methods}, 1(2):150.

\bibitem[{Caselli and Quochi(2007)}]{caselli2007inferring}
Tommaso Caselli and Valeria Quochi. 2007.
\newblock Inferring the semantics of temporal prepositions in italian.
\newblock In \emph{Proceedings of the Fourth ACL-SIGSEM Workshop on
  Prepositions}, pages 38--44. Association for Computational Linguistics.

\bibitem[{Che et~al.(2018)Che, Liu, Wang, Zheng, and Liu}]{che-EtAl:2018:K18-2}
Wanxiang Che, Yijia Liu, Yuxuan Wang, Bo~Zheng, and Ting Liu. 2018.
\newblock \href {http://www.aclweb.org/anthology/K18-2005} {Towards better {UD}
  parsing: Deep contextualized word embeddings, ensemble, and treebank
  concatenation}.
\newblock In \emph{Proceedings of the {CoNLL} 2018 Shared Task: Multilingual
  Parsing from Raw Text to Universal Dependencies}, pages 55--64, Brussels,
  Belgium. Association for Computational Linguistics.

\bibitem[{Cohn(2013)}]{cohn2013visual}
Neil Cohn. 2013.
\newblock Visual narrative structure.
\newblock \emph{Cognitive science}, 37(3):413--452.

\bibitem[{Comrie(1976)}]{comrie1976aspect}
Bernard Comrie. 1976.
\newblock \emph{Aspect: An introduction to the study of verbal aspect and
  related problems}, volume~2.
\newblock Cambridge university press.

\bibitem[{Cumming et~al.(2017)Cumming, Greenberg, and
  Kelly}]{cumming2017conventions}
Samuel Cumming, Gabriel Greenberg, and Rory Kelly. 2017.
\newblock Conventions of viewpoint coherence in film.
\newblock \emph{Philosophers' Imprint}, 17(1):1--29.

\bibitem[{Devlin et~al.(2019)Devlin, Chang, Lee, and
  Toutanova}]{devlin2019bert}
Jacob Devlin, Ming-Wei Chang, Kenton Lee, and Kristina Toutanova. 2019.
\newblock \href {http://arxiv.org/abs/1810.04805} {Bert: Pre-training of deep
  bidirectional transformers for language understanding}.

\bibitem[{Dorr and Olsen(1997)}]{Dorr_1997}
Bonnie~J. Dorr and Mari~Broman Olsen. 1997.
\newblock Deriving verbal and compositonal lexical aspect for nlp applications.
\newblock In \emph{Proceedings of the 35th Annual Meeting of the Association
  for Computational Linguistics}, pages 151--158, Madrid, Spain. Association
  for Computational Linguistics.

\bibitem[{Faul et~al.(2014)Faul, Erdfelder, Lang, and Buchner}]{faul2014g}
F~Faul, E~Erdfelder, AG~Lang, and A~Buchner. 2014.
\newblock G* power: statistical power analyses for windows and mac.

\bibitem[{Filip(2004)}]{Filip_2004}
Hana Filip. 2004.
\newblock The telicity parameter revisited.
\newblock In \emph{Semantics and Linguistic Theory}, volume~14, pages 92--109.

\bibitem[{Forman and Scholz(2010)}]{Forman_2010}
George Forman and Martin Scholz. 2010.
\newblock \href {https://doi.org/10.1145/1882471.1882479} {Apples-to-apples in
  cross-validation studies: Pitfalls in classifier performance measurement}.
\newblock \emph{SIGKDD Explor. Newsl.}, 12(1):49–57.

\bibitem[{Grave et~al.(2018)Grave, Bojanowski, Gupta, Joulin, and
  Mikolov}]{Grave_2018}
Edouard Grave, Piotr Bojanowski, Prakhar Gupta, Armand Joulin, and Tomas
  Mikolov. 2018.
\newblock Learning word vectors for 157 languages.
\newblock In \emph{Proceedings of the International Conference on Language
  Resources and Evaluation (LREC 2018)}.

\bibitem[{Gu{\'e}ron(2008)}]{Gueron_2008}
Jacqueline Gu{\'e}ron. 2008.
\newblock On the difference between telicity and perfectivity.
\newblock \emph{Lingua}, 118(11):1816--1840.

\bibitem[{Guillou et~al.(2020)Guillou, Bijl~de Vroe, Hosseini, Johnson, and
  Steedman}]{guillou-etal-2020-incorporating}
Liane Guillou, Sander Bijl~de Vroe, Mohammad~Javad Hosseini, Mark Johnson, and
  Mark Steedman. 2020.
\newblock \href {https://doi.org/10.18653/v1/2020.textgraphs-1.7}
  {Incorporating temporal information in entailment graph mining}.
\newblock In \emph{Proceedings of the Graph-based Methods for Natural Language
  Processing (TextGraphs)}, pages 60--71, Barcelona, Spain (Online).
  Association for Computational Linguistics.

\bibitem[{Habash(2010)}]{arabic-nlp-intro}
Nizar~Y. Habash. 2010.
\newblock \href {https://doi.org/10.2200/S00277ED1V01Y201008HLT010}
  {Introduction to arabic natural language processing}.
\newblock \emph{Synthesis Lectures on Human Language Technologies},
  3(1):1--187.

\bibitem[{Hamm and Bott(2018)}]{sep-tense-aspect}
Friedrich Hamm and Oliver Bott. 2018.
\newblock {Tense and Aspect}.
\newblock In Edward~N. Zalta, editor, \emph{The {Stanford} Encyclopedia of
  Philosophy}, fall 2018 edition. Metaphysics Research Lab, Stanford
  University.

\bibitem[{Jas and Parikh(2015)}]{jas2015image}
Mainak Jas and Devi Parikh. 2015.
\newblock Image specificity.
\newblock In \emph{Proceedings of the IEEE Conference on Computer Vision and
  Pattern Recognition}, pages 2727--2736.

\bibitem[{Kober et~al.(2020)Kober, Alikhani, Stone, and
  Steedman}]{kober-etal-2020-aspectuality}
Thomas Kober, Malihe Alikhani, Matthew Stone, and Mark Steedman. 2020.
\newblock \href {https://doi.org/10.18653/v1/2020.coling-main.401}
  {Aspectuality across genre: A distributional semantics approach}.
\newblock In \emph{Proceedings of the 28th International Conference on
  Computational Linguistics}, pages 4546--4562, Barcelona, Spain (Online).
  International Committee on Computational Linguistics.

\bibitem[{Kober et~al.(2019)Kober, Bijl~de Vroe, and
  Steedman}]{kober-etal-2019-temporal}
Thomas Kober, Sander Bijl~de Vroe, and Mark Steedman. 2019.
\newblock \href {https://doi.org/10.18653/v1/W19-0409} {Temporal and aspectual
  entailment}.
\newblock In \emph{Proceedings of the 13th International Conference on
  Computational Semantics - Long Papers}, pages 103--119, Gothenburg, Sweden.
  Association for Computational Linguistics.

\bibitem[{Korobov(2015)}]{Korobov_2015}
Mikhail Korobov. 2015.
\newblock \href {https://doi.org/10.1007/978-3-319-26123-2_31} {Morphological
  analyzer and generator for russian and ukrainian languages}.
\newblock In Mikhail~Yu. Khachay, Natalia Konstantinova, Alexander Panchenko,
  Dmitry~I. Ignatov, and Valeri~G. Labunets, editors, \emph{Analysis of Images,
  Social Networks and Texts}, volume 542 of \emph{Communications in Computer
  and Information Science}, pages 320--332. Springer International Publishing.

\bibitem[{Laleko(2008)}]{Laleko_2008}
Oksana Laleko. 2008.
\newblock Compositional telicity and heritage russian aspect.
\newblock In \emph{Proceedings of the Thirty-Eighth Western Conference on
  Linguistics (WECOL)}, volume~19, pages 150--160.

\bibitem[{Li et~al.(2019)Li, Tao, Li, and Fu}]{li2019visual}
Sheng Li, Zhiqiang Tao, Kang Li, and Yun Fu. 2019.
\newblock Visual to text: Survey of image and video captioning.
\newblock \emph{IEEE Transactions on Emerging Topics in Computational
  Intelligence}, 3(4):297--312.

\bibitem[{Lin(2006)}]{lin2006time}
Jo-Wang Lin. 2006.
\newblock Time in a language without tense: The case of chinese.
\newblock \emph{Journal of Semantics}, 23(1):1--53.

\bibitem[{Lin et~al.(2014)Lin, Maire, Belongie, Hays, Perona, Ramanan,
  Doll{\'a}r, and Zitnick}]{lin2014microsoft}
Tsung-Yi Lin, Michael Maire, Serge Belongie, James Hays, Pietro Perona, Deva
  Ramanan, Piotr Doll{\'a}r, and C~Lawrence Zitnick. 2014.
\newblock Microsoft coco: Common objects in context.
\newblock In \emph{European conference on computer vision}, pages 740--755.
  Springer.

\bibitem[{Majidi(2011)}]{Nemood}
Maryam Majidi. 2011.
\newblock Lexical aspect in farsi.
\newblock In \emph{Proceedings of the journal of Persian Literature}, pages
  145--158.

\bibitem[{McCloud(1993)}]{mccloud1993understanding}
Scott McCloud. 1993.
\newblock \emph{Understanding comics: The invisible art}.
\newblock William Morrow.

\bibitem[{Megerdoomian(2002)}]{megerdoomian2002aspect}
Karine Megerdoomian. 2002.
\newblock Aspect in complex predicates.
\newblock In \emph{Talk presented at the Workshop on Complex Predicates,
  Particles and Subevents, Konstanz}.

\bibitem[{Metheniti et~al.(2022)Metheniti, Van De~Cruys, and
  Hathout}]{metheniti-etal-2022-time}
Eleni Metheniti, Tim Van De~Cruys, and Nabil Hathout. 2022.
\newblock \href {https://doi.org/10.18653/v1/2022.cmcl-1.10} {About time: Do
  transformers learn temporal verbal aspect?}
\newblock In \emph{Proceedings of the Workshop on Cognitive Modeling and
  Computational Linguistics}, pages 88--101, Dublin, Ireland. Association for
  Computational Linguistics.

\bibitem[{Mirza and Tonelli(2014)}]{mirza-tonelli-2014-analysis}
Paramita Mirza and Sara Tonelli. 2014.
\newblock \href {https://aclanthology.org/C14-1198} {An analysis of causality
  between events and its relation to temporal information}.
\newblock In \emph{Proceedings of {COLING} 2014, the 25th International
  Conference on Computational Linguistics: Technical Papers}, pages 2097--2106,
  Dublin, Ireland. Dublin City University and Association for Computational
  Linguistics.

\bibitem[{Moens and Steedman(1988)}]{Moens_1988}
Marc Moens and Mark Steedman. 1988.
\newblock \href {https://www.aclweb.org/anthology/J88-2003} {Temporal ontology
  and temporal reference}.
\newblock \emph{Computational Linguistics}, 14(2):15--28.

\bibitem[{Molnar(2022)}]{molnar2022}
Christoph Molnar. 2022.
\newblock \href {https://christophm.github.io/interpretable-ml-book}
  {\emph{Interpretable Machine Learning}}, 2 edition.

\bibitem[{Oflazer et~al.(1994)Oflazer, Göçmen, Gocmen, and
  Bozsahin}]{Oflazer94anoutline}
Kemal Oflazer, Elvan Göçmen, Elvan Gocmen, and Cem Bozsahin. 1994.
\newblock \href
  {http://citeseerx.ist.psu.edu/viewdoc/download?doi=10.1.1.57.6951&rep=rep1&type=pdf}
  {An outline of turkish morphology}.

\bibitem[{Pasha et~al.(2014)Pasha, Al-Badrashiny, Diab, El~Kholy, Eskander,
  Habash, Pooleery, Rambow, and Roth}]{pasha-etal-2014-madamira}
Arfath Pasha, Mohamed Al-Badrashiny, Mona Diab, Ahmed El~Kholy, Ramy Eskander,
  Nizar Habash, Manoj Pooleery, Owen Rambow, and Ryan Roth. 2014.
\newblock \href
  {http://www.lrec-conf.org/proceedings/lrec2014/pdf/593_Paper.pdf}
  {{MADAMIRA}: A fast, comprehensive tool for morphological analysis and
  disambiguation of {A}rabic}.
\newblock In \emph{Proceedings of the Ninth International Conference on
  Language Resources and Evaluation ({LREC}-2014)}, pages 1094--1101,
  Reykjavik, Iceland. European Languages Resources Association (ELRA).

\bibitem[{Pedregosa et~al.(2011)Pedregosa, Varoquaux, Gramfort, Michel,
  Thirion, Grisel, Blondel, Prettenhofer, Weiss, Dubourg, Vanderplas, Passos,
  Cournapeau, Brucher, Perrot, and Duchesnay}]{Pedregosa_2011}
Fabian Pedregosa, Ga\"{e}l Varoquaux, Alexandre Gramfort, Vincent Michel,
  Bertrand Thirion, Olivier Grisel, Mathieu Blondel, Peter Prettenhofer, Ron
  Weiss, Vincent Dubourg, Jake Vanderplas, Alexandre Passos, David Cournapeau,
  Matthieu Brucher, Matthieu Perrot, and \'{E}douard Duchesnay. 2011.
\newblock \href {http://dl.acm.org/citation.cfm?id=1953048.2078195}
  {Scikit-learn: Machine learning in python}.
\newblock \emph{Journal of Machine Learning Research}, 12:2825--2830.

\bibitem[{Peters et~al.(2018)Peters, Neumann, Iyyer, Gardner, Clark, Lee, and
  Zettlemoyer}]{peters-etal-2018-deep}
Matthew~E. Peters, Mark Neumann, Mohit Iyyer, Matt Gardner, Christopher Clark,
  Kenton Lee, and Luke Zettlemoyer. 2018.
\newblock \href {https://doi.org/10.18653/v1/N18-1202} {Deep contextualized
  word representations}.
\newblock In \emph{Proceedings of the 2018 Conference of the North {A}merican
  Chapter of the Association for Computational Linguistics: Human Language
  Technologies, Volume 1 (Long Papers)}, pages 2227--2237, New Orleans,
  Louisiana. Association for Computational Linguistics.

\bibitem[{Pustejovsky et~al.(2010)Pustejovsky, Lee, Bunt, and
  Romary}]{pustejovsky2010iso}
James Pustejovsky, Kiyong Lee, Harry Bunt, and Laurent Romary. 2010.
\newblock Iso-timeml: An international standard for semantic annotation.
\newblock In \emph{LREC}, volume~10, pages 394--397.

\bibitem[{Schuler and Palmer(2005)}]{Schuler_2005}
Karin~Kipper Schuler and Martha~S. Palmer. 2005.
\newblock \emph{Verbnet: A Broad-Coverage, Comprehensive Verb Lexicon}.
\newblock Ph.D. thesis, USA.
\newblock AAI3179808.

\bibitem[{Shapley(1953)}]{Shapely_1953}
L.~S. Shapley. 1953.
\newblock \emph{A Value for n-Person Games}, pages 307--317. Princeton
  University Press.

\bibitem[{Slobin(2004)}]{slobin2004many}
Dan~I Slobin. 2004.
\newblock The many ways to search for a frog.
\newblock \emph{Relating events in narrative}, 2:219--257.

\bibitem[{Spencer(1991)}]{spencer1991morphological}
Andrew Spencer. 1991.
\newblock \emph{Morphological theory: An introduction to word structure in
  generative grammar}.
\newblock Wiley-Blackwell.

\bibitem[{Straka and Strakov{\'a}(2017)}]{Straka_2017}
Milan Straka and Jana Strakov{\'a}. 2017.
\newblock \href {https://doi.org/10.18653/v1/K17-3009} {Tokenizing, {POS}
  tagging, lemmatizing and parsing {UD} 2.0 with {UDP}ipe}.
\newblock In \emph{Proceedings of the {C}o{NLL} 2017 Shared Task: Multilingual
  Parsing from Raw Text to Universal Dependencies}, pages 88--99, Vancouver,
  Canada. Association for Computational Linguistics.

\bibitem[{Sun(2012)}]{sun2012jieba}
Junyi Sun. 2012.
\newblock Jieba.
\newblock \emph{Chinese word segmentation tool}.

\bibitem[{van Miltenburg et~al.(2018{\natexlab{a}})van Miltenburg,
  K{\'a}d{\'a}r, Koolen, and Krahmer}]{van-miltenburg-etal-2018-didec}
Emiel van Miltenburg, {\'A}kos K{\'a}d{\'a}r, Ruud Koolen, and Emiel Krahmer.
  2018{\natexlab{a}}.
\newblock \href {https://www.aclweb.org/anthology/C18-1310} {{DIDEC}: The
  {D}utch image description and eye-tracking corpus}.
\newblock In \emph{Proceedings of the 27th International Conference on
  Computational Linguistics}, pages 3658--3669, Santa Fe, New Mexico, USA.
  Association for Computational Linguistics.

\bibitem[{van Miltenburg et~al.(2018{\natexlab{b}})van Miltenburg, Koolen, and
  Krahmer}]{van-miltenburg-etal-2018-varying}
Emiel van Miltenburg, Ruud Koolen, and Emiel Krahmer. 2018{\natexlab{b}}.
\newblock \href {https://www.aclweb.org/anthology/W18-3910} {Varying image
  description tasks: spoken versus written descriptions}.
\newblock In \emph{Proceedings of the Fifth Workshop on {NLP} for Similar
  Languages, Varieties and Dialects ({V}ar{D}ial 2018)}, pages 88--100, Santa
  Fe, New Mexico, USA. Association for Computational Linguistics.

\bibitem[{Vinyals et~al.(2015)Vinyals, Toshev, Bengio, and
  Erhan}]{vinyals2015show}
Oriol Vinyals, Alexander Toshev, Samy Bengio, and Dumitru Erhan. 2015.
\newblock Show and tell: A neural image caption generator.
\newblock In \emph{Proceedings of the IEEE conference on computer vision and
  pattern recognition}, pages 3156--3164.

\bibitem[{Young et~al.(2014)Young, Lai, Hodosh, and
  Hockenmaier}]{young2014image}
Peter Young, Alice Lai, Micah Hodosh, and Julia Hockenmaier. 2014.
\newblock From image descriptions to visual denotations: New similarity metrics
  for semantic inference over event descriptions.
\newblock \emph{Transactions of the Association for Computational Linguistics},
  2:67--78.

\bibitem[{Çakmak(2013)}]{cakmak2013}
Serkan Çakmak. 2013.
\newblock \href
  {http://www.acarindex.com/dosyalar/makale/acarindex-1423932916.pdf} {"var" ve
  "yok" sözcüklerinin morfolojik kimliği.}
\newblock \emph{International Periodical For The Languages, Literature and
  History of Turkish or Turkic}, 8(4):463--471.

\end{thebibliography}
